\documentclass[a4paper, 11pt]{article} 

\usepackage[protrusion=true,expansion=true]{microtype} 
\usepackage{graphicx} 
\usepackage{wrapfig} 
\usepackage{color}
\usepackage{multicol}
\usepackage{hyperref}
\usepackage{amssymb,amsthm}
\usepackage{mathtools}

\usepackage{mathpazo} 
\usepackage[T1]{fontenc} 
\makeatletter

\renewcommand\@biblabel[1]{\textbf{#1.}} 
\renewcommand{\@listI}{\itemsep=0pt} 

\renewcommand{\maketitle}{ 
\begin{center} 
{\LARGE\@title} 

\vspace{50pt} 

{\large\@author} 
\\\@date 

\vspace{40pt} 
\end{center}
}

\title{\textbf{Tutorial on Variational Autoencoders}} 

\author{\textsc{Carl Doersch} 
\\{\textit{Carnegie Mellon / UC Berkeley}}} 

\date{August 16, 2016, with very minor revisions on January 3, 2021}

\begin{document}

\maketitle 



\begin{abstract}
In just three years, Variational Autoencoders (VAEs) have emerged as one of the most popular approaches to unsupervised learning of complicated distributions.  
VAEs are appealing because they are built on top of standard function approximators (neural networks), and can be trained with stochastic gradient descent.
VAEs have already shown promise in generating many kinds of complicated data, including handwritten digits~\cite{Kingma14a, Salimans15}, faces~\cite{Kingma14a,  Rezende14, kulkarni2015deep}, house numbers~\cite{Kingma14b, Gregor15}, CIFAR images~\cite{Gregor15}, physical models of scenes~\cite{kulkarni2015deep}, segmentation~\cite{sohn2015learning}, and predicting the future from static images~\cite{walker2016uncertain}. 
This tutorial introduces the intuitions behind VAEs, explains the mathematics behind them, and describes some empirical behavior.
No prior knowledge of variational Bayesian methods is assumed.  
\end{abstract}

\hspace*{3,6mm}\textit{Keywords:} variational autoencoders, unsupervised learning, structured prediction, neural networks

\vspace{30pt} 


\section{Introduction}
``Generative modeling'' is a broad area of machine learning which deals with models of distributions $P(X)$, defined over datapoints $X$ in some potentially high-dimensional space $\mathcal{X}$.  
For instance, images are a popular kind of data for which we might create generative models.
Each ``datapoint'' (image) has thousands or millions of dimensions (pixels), and the generative model's job is to somehow capture the dependencies between pixels, e.g., that nearby pixels have similar color, and are organized into objects.
Exactly what it means to ``capture'' these dependencies depends on exactly what we want to do with the model.
One straightforward kind of generative model simply allows us to compute $P(X)$ numerically.
In the case of images, $X$ values which look like real images should get high probability, whereas images that look like random noise should get low probability.
However, models like this are not necessarily useful: knowing that one image is unlikely does not help us synthesize one that is likely.

Instead, one often cares about producing more examples that are \textit{like} those already in a database, but not exactly the same.  
We could start with a database of raw images and synthesize new, unseen images.  
We might take in a database of 3D models of something like plants and produce more of them to fill a forest in a video game.
We could take handwritten text and try to produce more handwritten text.
Tools like this might actually be useful for graphic designers.
We can formalize this setup by saying that we get examples $X$ distributed according to some unknown distribution $P_{gt}(X)$, and our goal is to learn a model $P$ which we can sample from, such that $P$ is as similar as possible to $P_{gt}$.  

Training this type of model has been a long-standing problem in the machine learning community, and classically, most approaches have had one of three serious drawbacks.
First, they might require strong assumptions about the structure in the data.
Second, they might make severe approximations, leading to sub-optimal models.
Or third, they might rely on computationally expensive inference procedures like Markov Chain Monte Carlo.
More recently, some works have made tremendous progress in training neural networks as powerful function approximators through backpropagation~\cite{krizhevsky2012imagenet}.
These advances have given rise to promising frameworks which can use backpropagation-based function approximators to build generative models.

One of the most popular such frameworks is the Variational Autoencoder~\cite{Kingma14a,Rezende14}, the subject of this tutorial.
The assumptions of this model are weak, and training is fast via backpropagation.
VAEs do make an approximation, but the error introduced by this approximation is arguably small given high-capacity models.
These characteristics have contributed to a quick rise in their popularity. 

This tutorial is intended to be an informal introduction to VAEs, and not a formal scientific paper about them.
It is aimed at people who might have uses for generative models, but might not have a strong background in the variatonal Bayesian methods and ``minimum description length'' coding models on which VAEs are based.
This tutorial began its life as a presentation for computer vision reading groups at UC Berkeley and Carnegie Mellon, and hence has a bias toward a vision audience.
Suggestions for improvement are appreciated.

\subsection{Preliminaries: Latent Variable Models}

When training a generative model, the more complicated the dependencies between the dimensions, the more difficult the models are to train.  
Take, for example, the problem of generating images of handwritten characters.  
Say for simplicity that we only care about modeling the digits 0-9.  
If the left half of the character contains the left half of a 5, then the right half cannot contain the left half of a 0, or the character will very clearly not look like any real digit.  
Intuitively, it helps if the model first decides which character to generate before it assigns a value to any specific pixel.
This kind of decision is formally called a \textit{latent variable}.
That is, before our model draws anything, it first randomly samples a digit value $z$ from the set $[0,...,9]$, and then makes sure all the strokes match that character.  
$z$ is called `latent' because given just a character produced by the model, we don't necessarily know which settings of the latent variables generated the character.
We would need to infer it using something like computer vision.

Before we can say that our model is representative of our dataset, we need to make sure that for every datapoint $X$ in the dataset, there is one (or many) settings of the latent variables which causes the model to generate something very similar to $X$.  
Formally, say we have a vector of latent variables $z$ in a high-dimensional space $\mathcal{Z}$ which we can easily sample according to some probability density function (PDF) $P(z)$ defined over $\mathcal{Z}$.
Then, say we have a family of deterministic functions $f(z;\theta)$, parameterized by a vector $\theta$ in some space $\Theta$, where $f: \mathcal{Z}\times\Theta \to \mathcal{X}$.
$f$ is deterministic, but if $z$ is random and $\theta$ is fixed, then $f(z;\theta)$ is a random variable in the space $\mathcal{X}$.
We wish to optimize $\theta$ such that we can sample $z$ from $P(z)$ and, with high probability, $f(z;\theta)$ will be like the $X$'s in our dataset. 
 
To make this notion precise mathematically, we are aiming maximize the probability of each $X$ in the training set under the entire generative process, according to:

\begin{equation}
P(X)=\int P(X|z;\theta)P(z) dz.
\label{eq:total}
\end{equation}

\noindent Here, $f(z;\theta)$ has been replaced by a distribution $P(X|z;\theta)$, which allows us to make the dependence of $X$ on $z$ explicit by using the law of total probability.
The intuition behind this framework---called ``maximum likelihood''---is that if the model is likely to produce training set samples, then it is also likely to produce similar samples, and unlikely to produce dissimilar ones.  
In VAEs, the choice of this output distribution is often Gaussian, i.e., $P(X|z;\theta) = \mathcal{N}(X|f(z;\theta),\sigma^2*I)$.
That is, it has mean $f(z;\theta)$ and covariance equal to the identity matrix $I$ times some scalar $\sigma$ (which is a hyperparameter).
This replacement is necessary to formalize the intuition that some $z$ needs to result in samples that are merely \textit{like} $X$. 
In general, and particularly early in training, our model will not produce outputs that are identical to any particular $X$.  
By having a Gaussian distribution, we can use gradient descent (or any other optimization technique) to increase $P(X)$ by making $f(z;\theta)$ approach $X$ for some $z$, i.e., gradually making the training data more likely under the generative model.
This wouldn't be possible if $P(X|z)$ was a Dirac delta function, as it would be if we used $X=f(z;\theta)$ deterministically! 
Note that the output distribution is not \textit{required} to be Gaussian: for instance, if $X$ is binary, then $P(X|z)$ might be a Bernoulli parameterized by $f(z;\theta)$.
The important property is simply that $P(X|z)$ can be computed, and is continuous in $\theta$.
From here onward, we will omit $\theta$ from $f(z;\theta)$ to avoid clutter.

\begin{figure*}
\begin{center}
\includegraphics[width=0.5\textwidth]{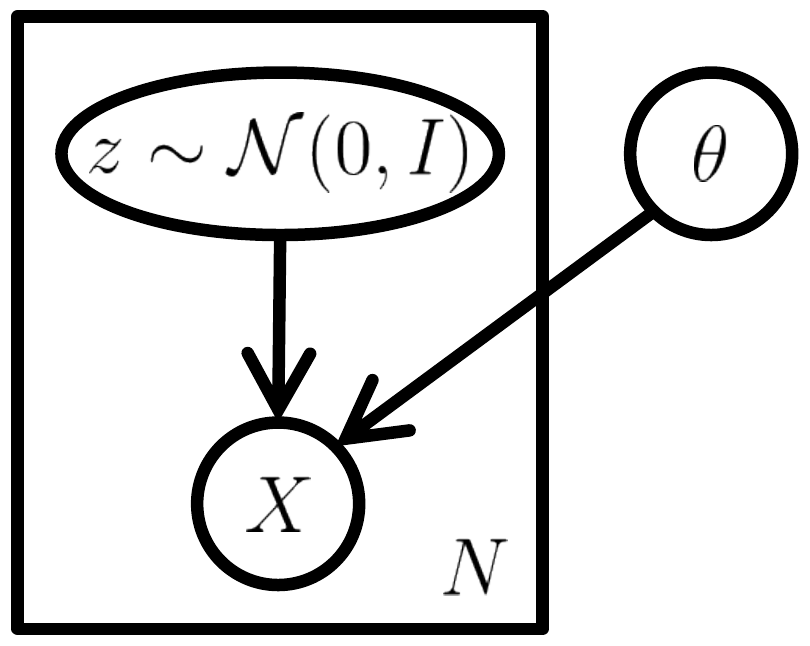}
\end{center}
\caption{The standard VAE model represented as a graphical model.  
Note the conspicuous lack of any structure or even an ``encoder'' pathway: it is possible to sample from the model without any input.
Here, the rectangle is ``plate notation'' meaning that we can sample from $z$ and $X$ $N$ times while the model parameters $\theta$ remain fixed.}
\label{fig:model}
\end{figure*}

\section{Variational Autoencoders}
\label{sec:vaes}
The mathematical basis of VAEs actually has relatively little to do with classical autoencoders, e.g. sparse autoencoders~\cite{olshausen1996emergence,lee2006efficient} or denoising autoencoders~\cite{vincent2008extracting,bengio2013deep}.  
VAEs approximately maximize Equation~\ref{eq:total}, according to the model shown in Figure~\ref{fig:model}.  
They are called ``autoencoders'' only because the final training objective that derives from this setup does have an encoder and a decoder, and \textit{resembles} a traditional autoencoder.  
Unlike sparse autoencoders, there are generally no tuning parameters analogous to the sparsity penalties.
And unlike sparse and denoising autoencoders, we can sample directly from $P(X)$ (without performing Markov Chain Monte Carlo, as in~\cite{bengio2013generalized}).  

To solve Equation~\ref{eq:total}, there are two problems that VAEs must deal with: how to define the latent variables $z$ (i.e., decide what information they represent), and how to deal with the integral over $z$.  
VAEs give a definite answer to both.  

\begin{figure*}
\begin{center}
\begin{tabular}{ccc}
\includegraphics[width=0.47\textwidth]{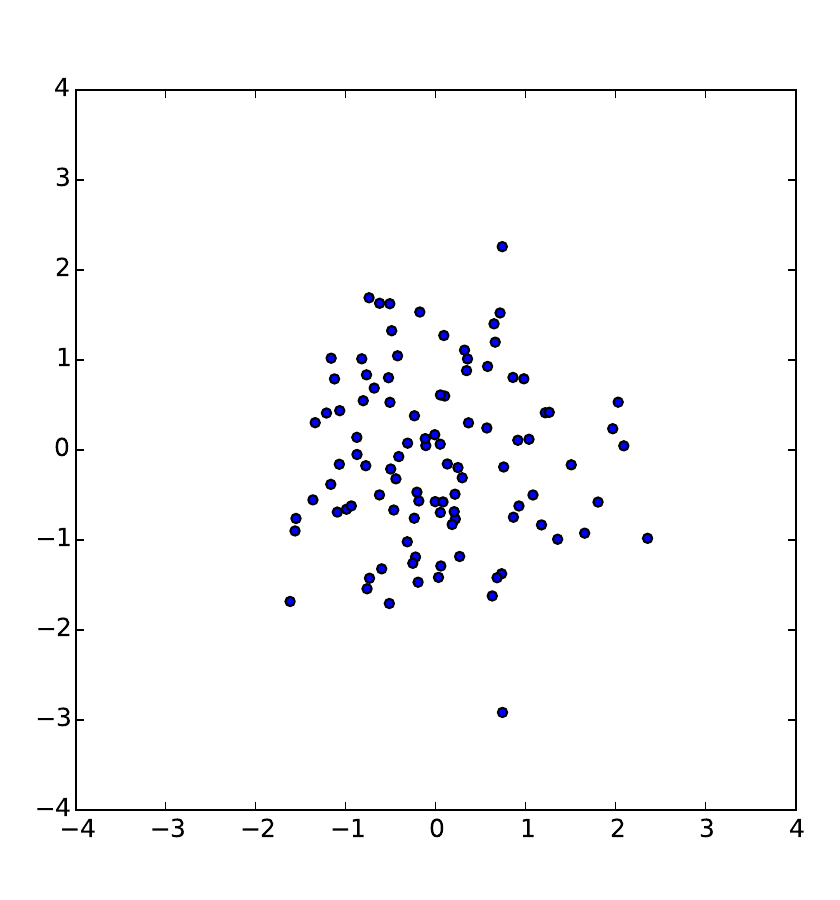} &
\includegraphics[width=0.47\textwidth]{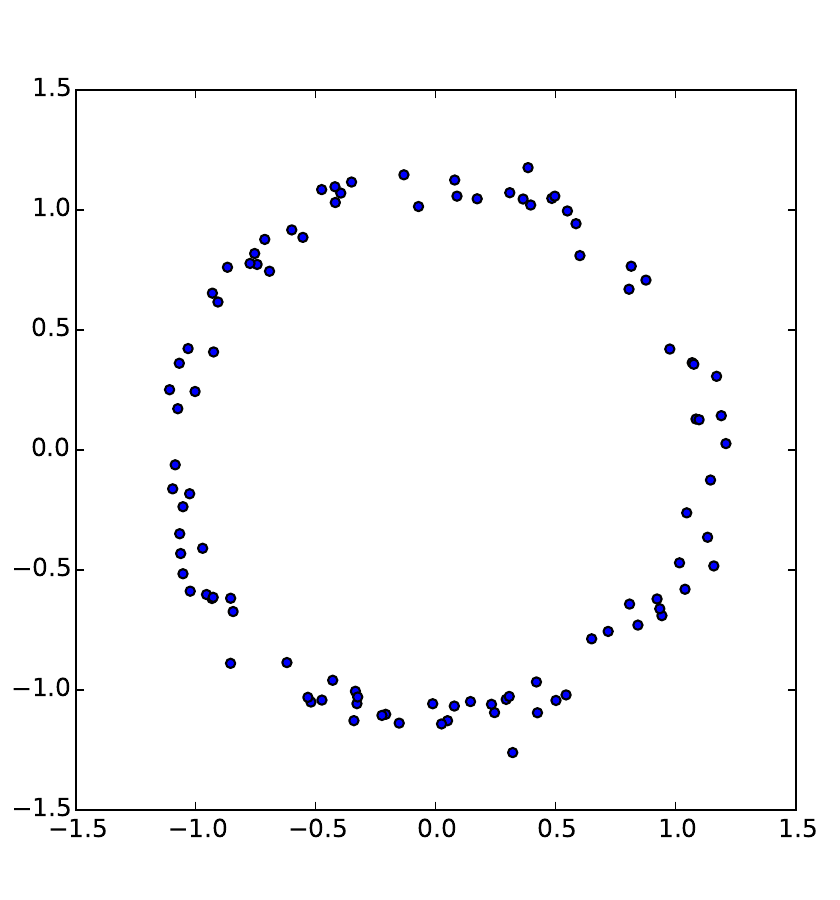} 
\end{tabular}
\end{center}
\caption{Given a random variable $z$ with one distribution, we can create another random variable $X=g(z)$ with a completely different distribution.
Left: samples from a gaussian distribution.
Right: those same samples mapped through the function $g(z)=z/10+z/||z||$ to form a ring.
This is the strategy that VAEs use to create arbitrary distributions: the deterministic function $g$ is learned from data.}
\label{fig:ring}
\end{figure*}

First, how do we choose the latent variables $z$ such that we capture latent information?  
Returning to our digits example, the `latent' decisions that the model needs to make before it begins painting the digit are actually rather complicated.  
It needs to choose not just the digit, but the angle that the digit is drawn, the stroke width, and also abstract stylistic properties.  
Worse, these properties may be correlated: a more angled digit may result if one writes faster, which also might tend to result in a thinner stroke.
Ideally, we want to avoid deciding by hand what information each dimension of $z$ encodes (although we may want to specify it by hand for some dimensions~\cite{kulkarni2015deep}).  
We also want to avoid explicitly describing the dependencies---i.e., the latent structure---between the dimensions of $z$.
VAEs take an unusual approach to dealing with this problem: they assume that there is no simple interpretation of the dimensions of $z$, and instead assert that samples of $z$ can be drawn from a simple distribution, i.e., $\mathcal{N}(0,I)$, where $I$ is the identity matrix.    
How is this possible?
The key is to notice that any distribution in $d$ dimensions can be generated by taking a set of $d$ variables that are normally distributed and mapping them through a sufficiently complicated function\footnote{In one dimension, you can use the inverse cumulative distribution function (CDF) of the desired distribution composed with the CDF of a Gaussian.  This is an extension of ``inverse transform sampling.''  For multiple dimensions, do the stated process starting with the marginal distribution for a single dimension, and repeat with the conditional distribution of each additional dimension.  See the ``inversion method'' and the ``conditional distribution method'' in Devroye et al.~\cite{devroye1986sample}}.  
For example, say we wanted to construct a 2D random variable whose values lie on a ring.  
If $z$ is 2D and normally distributed, $g(z)=z/10+z/||z||$ is roughly ring-shaped, as shown in Figure~\ref{fig:ring}.  
Hence, provided powerful function approximators, we can simply learn a function which maps our independent, normally-distributed $z$ values to whatever latent variables might be needed for the model, and then map those latent variables to $X$. 
In fact, recall that $P(X|z;\theta) = \mathcal{N}(X|f(z;\theta),\sigma^2*I)$.
If $f(z;\theta)$ is a multi-layer neural network, then we can imagine the network using its first few layers to map the normally distributed $z$'s to the latent values (like digit identity, stroke weight, angle, etc.) with exactly the right statistics.
Then it can use later layers to map those latent values to a fully-rendered digit.
In general, we don't need to worry about ensuring that the latent structure exists.
If such latent structure helps the model accurately reproduce (i.e. maximize the likelihood of) the training set, then the network will learn that structure at some layer.

\begin{figure*}
\begin{center}
\begin{tabular}{ccc}
\includegraphics[width=0.25\textwidth]{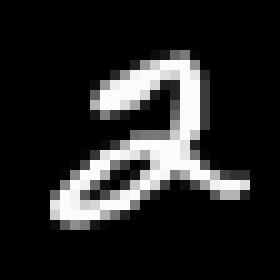} &
\includegraphics[width=0.25\textwidth]{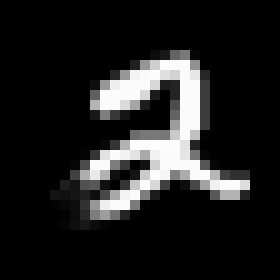} &
\includegraphics[width=0.25\textwidth]{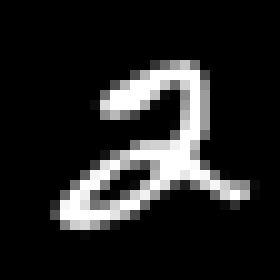} \\
(a) & (b) & (c)
\end{tabular}
\end{center}
\caption{It's hard to measure the likelihood of images under a model using only sampling.  
Given an image $X$ (a), the middle sample (b) is much closer in Euclidean distance than the one on the right (c).
Because pixel distance is so different from perceptual distance, a sample needs to be extremely close in pixel distance to a datapoint $X$ before it can be considered evidence that $X$ is likely under the model.}
\label{fig:digits}
\end{figure*}

Now all that remains is to maximize Equation~\ref{eq:total}, where $P(z)=\mathcal{N}(z|0,I)$.  
As is common in machine learning, if we can find a computable formula for $P(X)$, and we can take the gradient of that formula, then we can optimize the model using stochastic gradient ascent. 
It is actually conceptually straightforward to compute $P(X)$ approximately: we first sample a large number of $z$ values $\{ z_{1} , ..., z_{n}\}$, and compute $P(X)\approx\frac{1}{n}\sum_i P(X|z_i)$.  
The problem here is that in high dimensional spaces, $n$ might need to be extremely large before we have an accurate estimate of $P(X)$.  
To see why, consider our example of handwritten digits.  
Say that our digit datapoints are stored in pixel space, in 28x28 images as shown in Figure~\ref{fig:digits}.
Since $P(X|z)$ is an isotropic Gaussian, the negative log probability of $X$ is proportional squared Euclidean distance between $f(z)$ and $X$.
Say that Figure~\ref{fig:digits}(a) is the target ($X$) for which we are trying to find $P(X)$.
A model which produces the image shown in Figure~\ref{fig:digits}(b) is probably a bad model, since this digit is not much like a 2.
Hence, we should set the $\sigma$ hyperparameter of our Gaussian distribution such that this kind of erroneous digit does not contribute to $P(X)$.
On the other hand, a model which produces Figure~\ref{fig:digits}(c) (identical to $X$ but shifted down and to the right by half a pixel) might be a good model.
We would hope that this sample would contribute to  $P(X)$.
Unfortunately, however, we can't have it both ways: the squared distance between $X$ and Figure~\ref{fig:digits}(c) is .2693 (assuming pixels range between 0 and 1), but between $X$ and Figure~\ref{fig:digits}(b) it is just .0387.  
The lesson here is that in order to reject samples like Figure~\ref{fig:digits}(b), we need to set $\sigma$ very small, such that the model needs to generate something \textit{significantly} more like $X$ than Figure~\ref{fig:digits}(c)!
Even if our model is an accurate generator of digits, we would likely need to sample many thousands of digits before we produce a 2 that is sufficiently similar to the one in Figure~\ref{fig:digits}(a).  
We might solve this problem by using a better similarity metric, but in practice these are difficult to engineer in complex domains like vision, and they're difficult to train without labels that indicate which datapoints are similar to each other.
Instead, VAEs alter the sampling procedure to make it faster, without changing the similarity metric.

\subsection{Setting up the objective}

Is there a shortcut we can take when using sampling to compute Equation~\ref{eq:total}? 
In practice, for most $z$, $P(X|z)$ will be nearly zero, and hence contribute almost nothing to our estimate of $P(X)$.
The key idea behind the variational autoencoder is to attempt to sample values of $z$ that are likely to have produced $X$, and compute $P(X)$ just from those.
This means that we need a new function $Q(z|X)$ which can take a value of $X$ and give us a distribution over $z$ values that are likely to produce $X$.  
Hopefully the space of $z$ values that are likely under $Q$ will be much smaller than the space of all $z$'s that are likely under the prior $P(z)$.
This lets us, for example, compute $E_{z\sim Q}P(X|z)$ relatively easily.
However, if $z$ is sampled from an arbitrary distribution with PDF $Q(z)$, which is not $\mathcal{N}(0,I)$, then how does that help us optimize $P(X)$?
The first thing we need to do is relate $E_{z\sim Q}P(X|z)$ and $P(X)$.
We'll see where $Q$ comes from later.

The relationship between $E_{z\sim Q}P(X|z)$ and $P(X)$ is one of the cornerstones of variational Bayesian methods.
We begin with the definition of Kullback-Leibler divergence (KL divergence or $\mathcal{D}$) between $P(z|X)$ and $Q(z)$, for some arbitrary $Q$ (which may or may not depend on $X$):
\begin{equation}
    \mathcal{D}\left[Q(z)\|P(z|X)\right]=E_{z\sim Q}\left[\log Q(z) - \log P(z|X) \right].
\label{eq:kl}
\end{equation}
\noindent We can get both $P(X)$ and $P(X|z)$ into this equation by applying Bayes rule to $P(z|X)$:
\begin{equation}
    \mathcal{D}\left[Q(z)\|P(z|X)\right]=E_{z\sim Q}\left[\log Q(z) - \log P(X|z) - \log P(z) \right] + \log P(X).
\end{equation}
\noindent Here, $\log P(X)$ comes out of the expectation because it does not depend on $z$.  Negating both sides, rearranging, and contracting part of $E_{z\sim Q}$ into a KL-divergence terms yields:
\begin{equation}
    \log P(X) - \mathcal{D}\left[Q(z)\|P(z|X)\right]=E_{z\sim Q}\left[\log P(X|z)  \right] - \mathcal{D}\left[Q(z)\|P(z)\right].
\end{equation}
\noindent Note that $X$ is fixed, and $Q$ can be \textit{any} distribution, not just a distribution which does a good job mapping $X$ to the $z$'s that can produce $X$.  
Since we're interested in inferring $P(X)$, it makes sense to construct a $Q$ which \textit{does} depend on $X$, and in particular, one which makes $\mathcal{D}\left[Q(z)\|P(z|X)\right]$ small:
\begin{equation}
    \log P(X) - \mathcal{D}\left[Q(z|X)\|P(z|X)\right]=E_{z\sim Q}\left[\log P(X|z)  \right] - \mathcal{D}\left[Q(z|X)\|P(z)\right].
    \label{eq:variational}
\end{equation}
\noindent This equation serves as the core of the variational autoencoder, and it's worth spending some time thinking about what it says\footnote{
Historically, this math (particularly Equation~\ref{eq:variational}) was known long before VAEs.
For example, Helmholtz Machines~\cite{dayan1995helmholtz} (see Equation 5) use nearly identical mathematics, with one crucial difference.
The integral in our expectations is replaced with a sum in Dayan et al.~\cite{dayan1995helmholtz}, because Helmholtz Machines assume a discrete distribution for the latent variables.  
This choice prevents the transformations that make gradient descent tractable in VAEs.
}.  
In two sentences, the left hand side has the quantity we want to maximize: $\log P(X)$ (plus an error term, which makes $Q$ produce $z$'s that can reproduce a given $X$; this term will become small if $Q$ is high-capacity).
The right hand side is something we can optimize via stochastic gradient descent given the right choice of $Q$ (although it may not be obvious yet how).
Note that the framework---in particular, the right hand side of Equation~\ref{eq:variational}---has suddenly taken a form which looks like an autoencoder, since $Q$ is ``encoding'' $X$ into $z$, and $P$ is ``decoding'' it to reconstruct $X$.
We'll explore this connection in more detail later.

Now for a bit more detail on Equatinon~\ref{eq:variational}.
Starting with the left hand side, we are maximizing $\log P(X)$ while simultaneously minimizing $\mathcal{D}\left[Q(z|X)\|P(z|X)\right]$.  
$P(z|X)$ is not something we can compute analytically: it describes the values of $z$ that are likely to give rise to a sample like $X$ under our model in Figure~\ref{fig:model}.  
However, the second term on the left is pulling $Q(z|x)$ to match $P(z|X)$.
Assuming we use an arbitrarily high-capacity model for $Q(z|x)$, then $Q(z|x)$ will hopefully actually \textit{match} $P(z|X)$, in which case this KL-divergence term will be zero, and we will be directly optimizing $\log P(X)$.  
As an added bonus, we have made the intractable $P(z|X)$ tractable: we can just use $Q(z|x)$ to compute it.

\subsection{Optimizing the objective}

So how can we perform stochastic gradient descent on the right hand side of Equation~\ref{eq:variational}?
First we need to be a bit more specific about the form that $Q(z|X)$ will take.  
The usual choice is to say that $Q(z|X)=\mathcal{N}(z|\mu(X;\vartheta),\Sigma(X;\vartheta))$, where $\mu$ and $\Sigma$ are arbitrary deterministic functions with parameters $\vartheta$ that can be learned from data (we will omit $\vartheta$ in later equations).
In practice, $\mu$ and $\Sigma$ are again implemented via neural networks, and $\Sigma$ is constrained to be a diagonal matrix.
The advantages of this choice are computational, as they make it clear how to compute the right hand side.
The last term---$\mathcal{D}\left[Q(z|X)\|P(z)\right]$---is now a KL-divergence between two multivariate Gaussian distributions, which can be computed in closed form as:
\begin{equation}
\begin{array}{c}
 \mathcal{D}[\mathcal{N}(\mu_0,\Sigma_0) \| \mathcal{N}(\mu_1,\Sigma_1)] = \hspace{20em}\\
  \hspace{5em}\frac{ 1 }{ 2 } \left( \mathrm{tr} \left( \Sigma_1^{-1} \Sigma_0 \right) + \left( \mu_1 - \mu_0\right)^\top \Sigma_1^{-1} ( \mu_1 - \mu_0 ) - k + \log \left( \frac{ \det \Sigma_1 }{ \det \Sigma_0  } \right)  \right)
\end{array}
\end{equation}
\noindent where $k$ is the dimensionality of the distribution.  In our case, this simplifies to:
\begin{equation}
\begin{array}{c}
 \mathcal{D}[\mathcal{N}(\mu(X),\Sigma(X)) \| \mathcal{N}(0,I)] = \hspace{20em}\\
\hspace{6em}\frac{ 1 }{ 2 } \left( \mathrm{tr} \left( \Sigma(X) \right) + \left( \mu(X)\right)^\top ( \mu(X) ) - k - \log\det\left(  \Sigma(X)  \right)  \right).
\end{array}
\end{equation}

\begin{figure*}
\begin{center}
\includegraphics[width=0.98\textwidth]{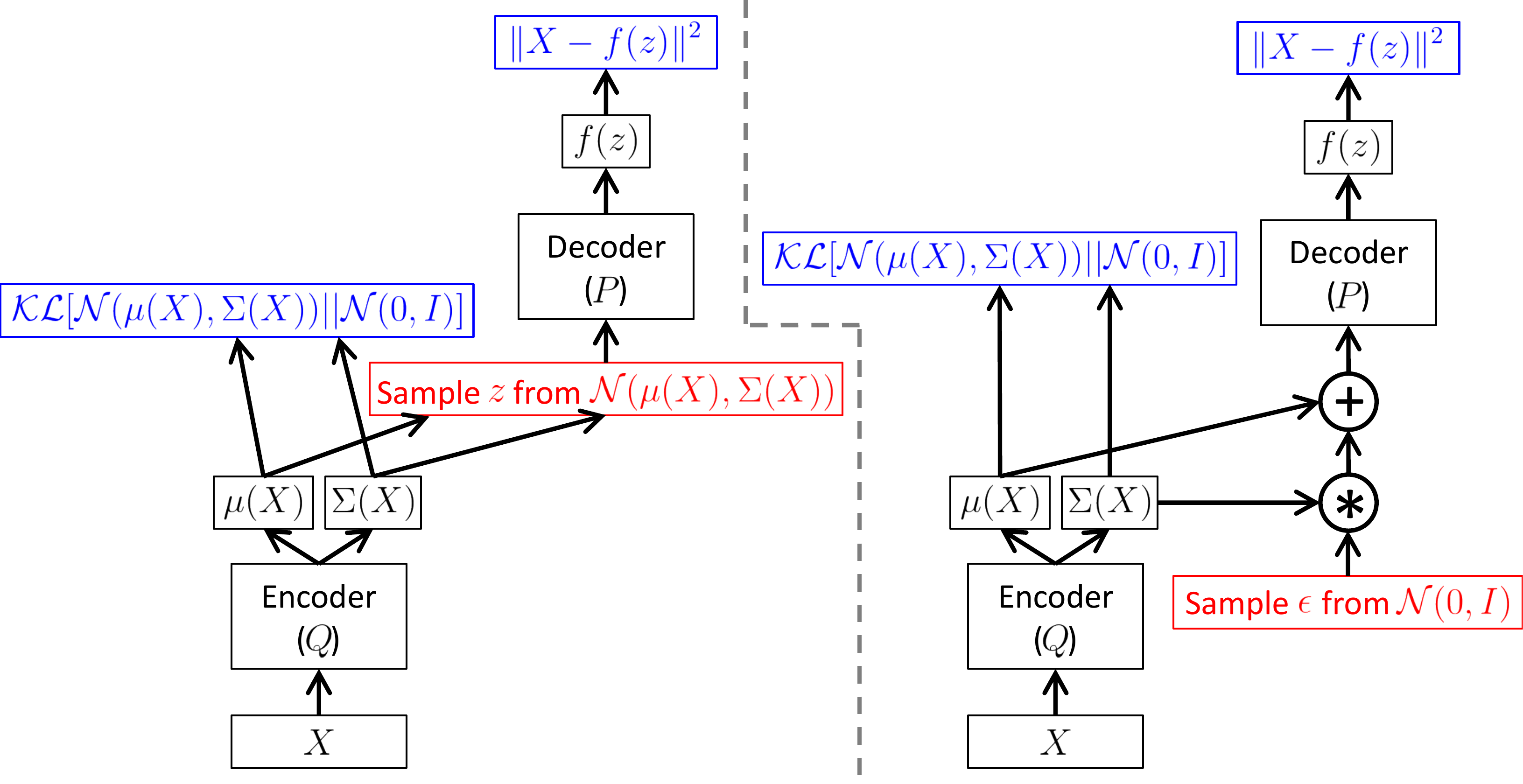} 
\end{center}
\caption{A training-time variational autoencoder implemented as a feedforward neural network, where $P(X|z)$ is Gaussian.  Left is without the ``reparameterization trick'', and right is with it.  Red shows sampling operations that are non-differentiable.  Blue shows loss layers.  The feedforward behavior of these networks is identical, but backpropagation can be applied only to the right network.}
\label{fig:net}
\end{figure*}

The first term on the right hand side of Equation~\ref{eq:variational} is a bit more tricky.
We could use sampling to estimate $E_{z\sim Q}\left[\log P(X|z)  \right]$, but getting a good estimate would require passing many samples of $z$ through $f$, which would be expensive.
Hence, as is standard in stochastic gradient descent, we take one sample of $z$ and treat $\log P(X|z)$ for that $z$ as an approximation of $E_{z\sim Q}\left[\log P(X|z)  \right]$.
After all, we are already doing stochastic gradient descent over different values of $X$ sampled from a dataset $D$.
The full equation we want to optimize is:
\begin{equation}
\begin{array}{c}
    E_{X\sim D}\left[\log P(X) - \mathcal{D}\left[Q(z|X)\|P(z|X)\right]\right]=\hspace{16em}\\
\hspace{10em}E_{X\sim D}\left[E_{z\sim Q}\left[\log P(X|z)  \right] - \mathcal{D}\left[Q(z|X)\|P(z)\right]\right].
\end{array}
    \label{eq:expected}
\end{equation}
If we take the gradient of this equation, the gradient symbol can be moved into the expectations.
Therefore, we can sample a single value of $X$ and a single value of $z$ from the distribution $Q(z|X)$, and compute the gradient of:
\begin{equation}
 \log P(X|z)-\mathcal{D}\left[Q(z|X)\|P(z)\right].
  \label{eq:onesamp}
\end{equation}
We can then average the gradient of this function over arbitrarily many samples of $X$ and $z$, and the result converges to the gradient of Equation~\ref{eq:expected}.

There is, however, a significant problem with Equation~\ref{eq:onesamp}.
$E_{z\sim Q}\left[\log P(X|z)  \right]$ depends not just on the parameters of $P$, but also on the parameters of $Q$.
However, in Equation~\ref{eq:onesamp}, this dependency has disappeared!
In order to make VAEs work, it's essential to drive $Q$ to produce codes for $X$ that $P$ can reliably decode.  
To see the problem a different way, the network described in Equation~\ref{eq:onesamp} is much like the network shown in Figure~\ref{fig:net} (left).
The forward pass of this network works fine and, if the output is averaged over many samples of $X$ and $z$, produces the correct expected value.
However, we need to back-propagate the error through a layer that samples $z$ from $Q(z|X)$, which is a non-continuous operation and has no gradient.
Stochastic gradient descent via backpropagation can handle stochastic inputs, but not stochastic units within the network!
The solution, called the ``reparameterization trick'' in~\cite{Kingma14a}, is to move the sampling to an input layer.
Given $\mu(X)$ and $\Sigma(X)$---the mean and covariance of $Q(z|X)$---we can sample from $\mathcal{N}(\mu(X),\Sigma(X))$ by first sampling $\epsilon \sim \mathcal{N}(0,I)$, then computing $z=\mu(X)+\Sigma^{1/2}(X)*\epsilon$. 
Thus, the equation we actually take the gradient of is:
\begin{equation}
 E_{X\sim D}\left[E_{\epsilon\sim\mathcal{N}(0,I)}[\log P(X|z=\mu(X)+\Sigma^{1/2}(X)*\epsilon)]-\mathcal{D}\left[Q(z|X)\|P(z)\right]\right].
\end{equation}
This is shown schematically in Figure~\ref{fig:net} (right).  
Note that none of the expectations are with respect to distributions that depend on our model parameters, so we can safely move a gradient symbol into them while maintaning equality.
That is, given a fixed $X$ and $\epsilon$, this function is deterministic and continuous in the parameters of $P$ and $Q$, meaning backpropagation can compute a gradient that will work for stochastic gradient descent.
It's worth pointing out that the ``reparameterization trick'' only works if we can sample from $Q(z|X)$ by evaluating a function $h(\eta,X)$, where $\eta$ is noise from a distribution that is not learned.
Furthermore, $h$ must be \textit{continuous} in $X$ so that we can backprop through it.  
This means $Q(z|X)$ (and therefore $P(z)$) can't be a discrete distribution!
If $Q$ is discrete, then for a fixed $\eta$, either $h$ needs to ignore $X$, or there needs to be some point at which $h(\eta,X)$ ``jumps'' from one possible value in $Q$'s sample space to another, i.e., a discontinuity.

\begin{figure*}
\begin{center}
\includegraphics[width=0.3\textwidth]{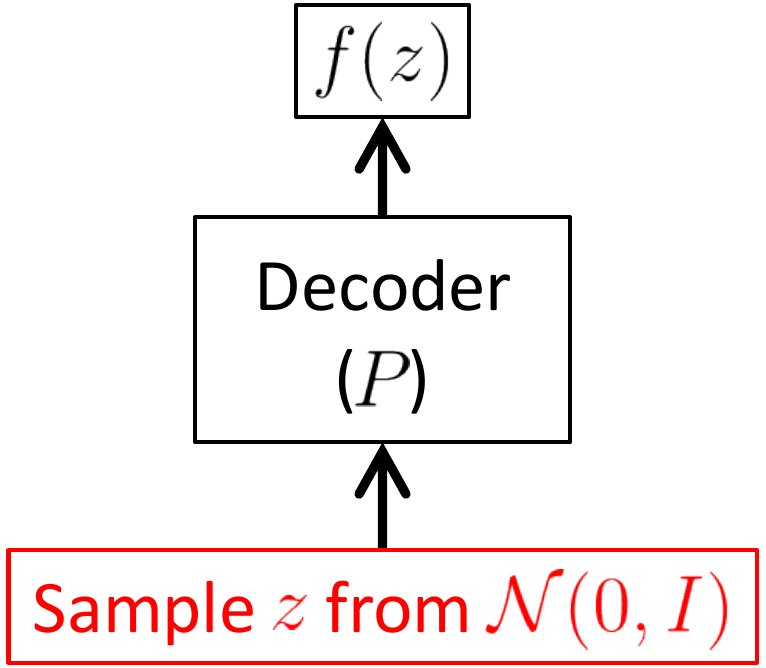} 
\end{center}
\caption{The testing-time variational ``autoencoder,'' which allows us to generate new samples.  The ``encoder'' pathway is simply discarded.}
\label{fig:test_time}
\end{figure*}

\subsection{Testing the learned model}

At test time, when we want to generate new samples, we simply input values of $z\sim\mathcal{N}(0,I)$ into the decoder.
That is, we remove the ``encoder,'' including the multiplication and addition operations that would change the distribution of $z$.
This (remarkably simple) test-time network is shown in Figure~\ref{fig:test_time}.

Say that we want to evaluate the probability of a testing example under the model.
This is, in general, not tractable.
Note, however, that $\mathcal{D}[Q(z|X)\|P(z|X)]$ is positive, meaning that the right hand side of Equation~\ref{eq:variational} is a lower bound to $P(X)$.  
This lower bound still can't quite be computed in closed form due to the expectation over $z$, which requires sampling.
However, sampling $z$ from $Q$ gives an estimator for the expectation which generally converges much faster than sampling $z$ from $\mathcal{N}(0,I)$ as discussed in section~\ref{sec:vaes}.  
Hence, this lower bound can be a useful tool for getting a rough idea of how well our model is capturing a particular datapoint $X$.

\subsection{Interpreting the objective}

By now, you are hopefully convinced that the learning in VAEs is tractable, and that it optimizes something like $\log P(X)$ across our entire dataset.
However, we are not optimizing \textit{exactly} $\log P(X)$, so this section aims to take a deeper look at what the objective function is actually doing.
We address three topics.
First, we ask how much error is introduced by optimizing $\mathcal{D}[Q(z|X)\|P(z|X)]$ in addition to $\log P(X)$.
Second, we describe the VAE framework---especially the r.h.s. of Equation~\ref{eq:variational}---in terms of information theory, linking it to other approaches based on Minimum Description Length.
Finally, we investigate whether VAEs have ``regularization parameters'' analogous to the sparsity penalty in sparse autoencoders.

\subsubsection{The error from $\mathcal{D}[Q(z|X)\|P(z|X)]$} The tractability of this model relies on our assumption that $Q(z|X)$ can be modeled as a Gaussian with some mean $\mu(X)$ and variance $\Sigma(X)$.  
$P(X)$ converges (in distribution) to the true distribution if and only if $\mathcal{D}[Q(z|X)\|P(z|X)]$ goes to zero.
Unfortunately, it's not straightforward to ensure that this happens. 
Even if we assume $\mu(X)$ and $\Sigma(X)$ are arbitrarily high capacity, the posterior $P(z|X)$ is not necessarily Gaussian for an arbitrary $f$ function that we're using to define $P$.
For fixed $P$, this might mean that $\mathcal{D}[Q(z|X)\|P(z|X)]$ never goes to zero.
However, the good news is that given sufficiently high-capacity neural networks, there are many $f$ functions that result in our model generating any given output distribution.
Any of these functions will maximize $\log P(X)$ equally well.
Hence, all we need is one function $f$ which both maximizes $\log P(X)$ \textit{and} results in $P(z|X)$ being Gaussian for all $X$.
If so, $\mathcal{D}[Q(z|X)\|P(z|X)]$ will pull our model towards that parameterization of the distribution.
So, does such a function exist for all distributions we might want to approximate?
I'm not aware of anyone proving this in general just yet, but it turns out that one can prove that such a function does exist, provided $\sigma$ is small relative to the curvature of the ground truth distribution's CDF (at least in 1D; a proof is included in Appendix~\ref{appendix:convergence}).
In practice such a small $\sigma$ might cause problems for existing machine learning algorithms, since the gradients would become badly scaled.  
However, it is comforting to know that VAEs have zero approximation error in at least this one scenario.
This fact suggests that future theoretical work may show us how much approximation error VAEs have in more practical setups.
(It seems to me like it should be possible to extend the proof technique in Appendix~\ref{appendix:convergence} to multiple dimensions, but this is left for future work.)

\subsubsection{The information-theoretic interpretation} Another important way to look at the right hand side of Equation~\ref{eq:variational} is in terms of information theory, and in particular, the ``minimum description length'' principle which motivated many of the VAE's predecessors like Helmholtz Machines~\cite{dayan1995helmholtz}, the Wake-Sleep Algorithm~\cite{hinton1995wake}, Deep Belief Nets~\cite{hinton2006fast}, and Boltzmann Machines~\cite{salakhutdinov2009deep}.
$-\log P(X)$ can be seen as the total number of bits required to construct a given $X$ under our model using an ideal encoding.
The right hand side of Equation~\ref{eq:variational} views this as a two-step process to construct $X$.
We first use some bits to construct $z$.
Recall that a KL-divergence is in units of bits (or, more precisely, nats).
Specifically, $\mathcal{D}[Q(z|X)||P(z)]$ is the expected information that's required to convert an uninformative sample from $P(z)$ into a sample from $Q(z|X)$ (the so-called ``information gain'' interpretation of KL-divergence).
That is, it measures the amount of \textit{extra} information that we get about $X$ when $z$ comes from $Q(z|X)$ instead of from $P(z)$ (for more details, see the ``bits back'' argument of~\cite{hinton1993keeping,hinton1994autoencoders}).
In the second step, $P(X|z)$ measures the amount of information required to reconstruct $X$ from $z$ under an ideal encoding.  
Hence, the total number of bits ($-\log P(X)$) is the sum of these two steps, minus a penalty we pay for $Q$ being a sub-optimal encoding ($\mathcal{D}[Q(z|X)||P(z|X)]$).

\subsubsection{VAEs and the regularization parameter} Looking at Equation~\ref{eq:variational}, it's interesting to view the $\mathcal{D}[Q(z|X)||P(z)]$ as a regularization term, much like the sparsity regularization in sparse autoencoders~\cite{olshausen1996emergence}.
From this standpoint, it's interesting to ask whether the variational autoencoder has any ``regularization parameter.''
That is, in the sparse autoencoder objective, we have a $\lambda$ regularization parameter in an objective function that looks something like this:
\begin{equation}
 \|\phi(\psi(X))-X\|^{2}+\lambda \|\psi(X)\|_0
\end{equation}
where $\psi$ and $\phi$ are the encoder and decoder functions, respectively, and $\|\cdot\|_0$ is an $L_{0}$ norm that encourages the encoding to be sparse.  
This $\lambda$ must be set by hand.

A variational autoencoder, however, does \textbf{not}, in general, have such a regularization parameter, which is good because that's one less parameter that the programmer needs to adjust.
However, for certain models, we can make it appear like such a regularization parameter exists.
It's tempting to think that this parameter can come from changing $z\sim\mathcal{N}(0,I)$ to something like $z^{\prime}\sim\mathcal{N}(0,\lambda*I)$, but it turns out that this doesn't change the model.
To see this, note that we can absorb this constant into $P$ and $Q$ by writing them in terms of $f^{\prime}(z^{\prime})=f(z^{\prime}/\lambda)$, $\mu^{\prime}(X)=\mu(X)*\lambda$, and $\Sigma^{\prime}(X)=\Sigma(X)*\lambda^2$.
This will produce an objective function whose value (right hand side of Equation~\ref{eq:variational}) is identical to the loss we had with $z\sim\mathcal{N}(0,I)$.
Also, the model for sampling $X$ will be identical, since $z^{\prime}/\lambda\sim\mathcal{N}(0,I)$.  

However, there is another place where a regularization parameter can come from.
Recall that a good choice for the output distribution for continuous data is $P(X|z)\sim\mathcal{N}(f(z),\sigma^2 * I)$ for some $\sigma$ we supply.
Therefore, $\log P(X|z)=C-\frac{1}{2}\|X-f(z)\|^{2}/\sigma^2$ (where $C$ is a constant that does not depend on $f$, and can be ignored during optimization).
When we write the full optimization objective, $\sigma$ appears in the second term on the r.h.s. of Equation~\ref{eq:variational}, but not the first; in this sense, the $\sigma$ we choose behaves like a $\lambda$ controlling the weighting of the two terms.
Note, however, that the existence of this parameter relies on our choice of the distribution of $X$ given $z$.
If $X$ is binary and we use a Bernoulli output model, then this regularization parameter disappears, and the only way to bring it back is to use hacks like replicating the dimensions of $X$.
From an information theory standpoint, this makes sense: when $X$ is binary, we can actually count the number of bits that are required to encode $X$, and both terms on the right hand side of Equation~\ref{eq:variational} are using the same units to count these bits.
However, when $X$ is continuous, each sample contains infinite information.
Our choice of $\sigma$ determines how accurately we expect the model to reconstruct $X$, which is necessary so that the information content can become finite.

\begin{figure*}
\begin{center}
\includegraphics[width=0.96\textwidth]{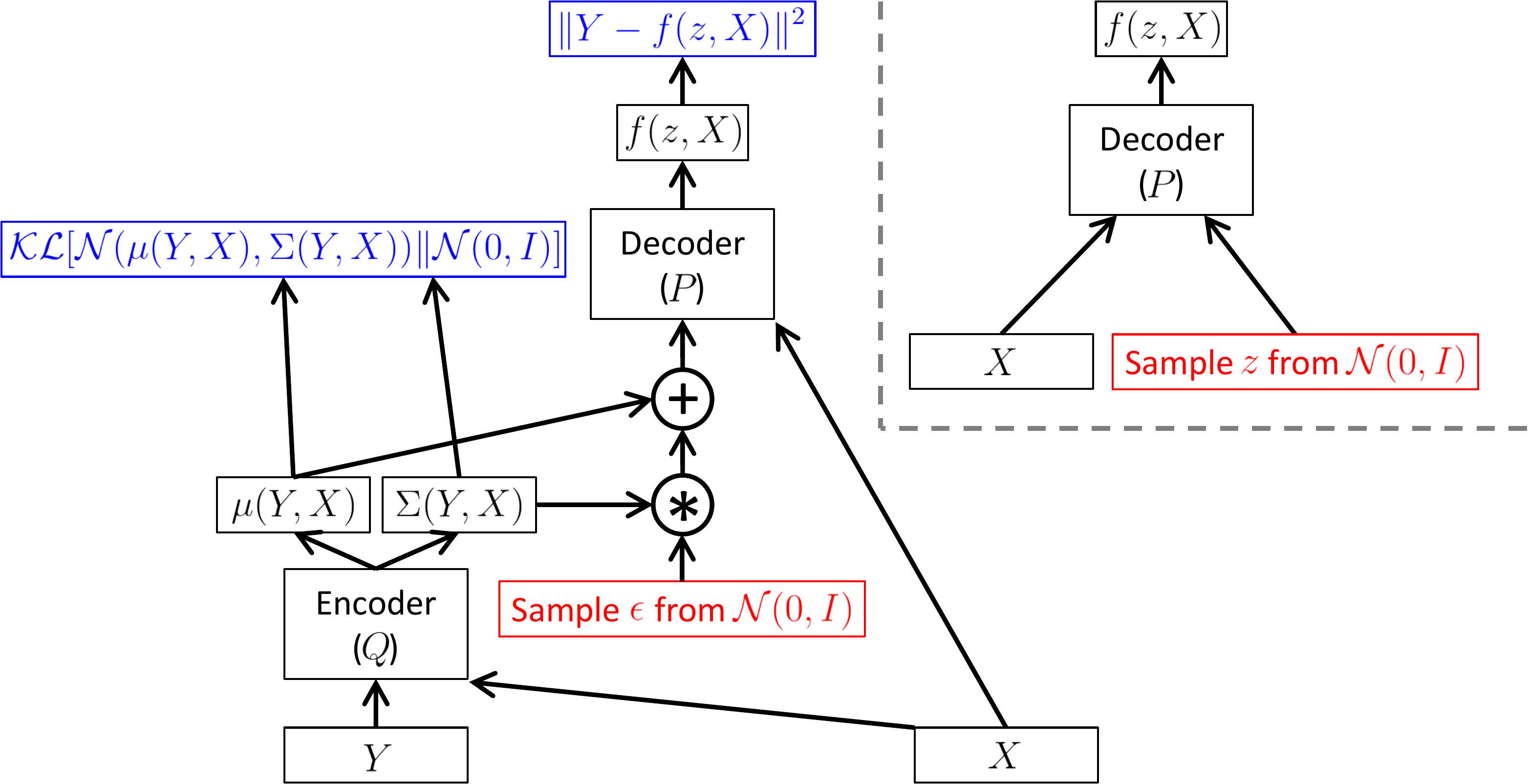} 
\end{center}
\caption{Left: a training-time conditional variational autoencoder implemented as a feedforward neural network, following the same notation as Figure~\ref{fig:net}. Right: the same model at test time, when we want to sample from $P(Y|X)$.}
\label{fig:cvae}
\end{figure*}

\section{Conditional Variational Autoencoders}

Let's return to our running example of generating handwritten digits.
Say that we don't just want to generate new digits, but instead we want to add digits to an existing string of digits written by a single person.
This is similar to a truly practical problem in computer graphics called hole filling: given an existing image where a user has removed an unwanted object, the goal is to fill in the hole with plausible-looking pixels.
An important difficulty with both problems is that the space of plausible outputs is \textit{multi-modal}: there are many possibilities for the next digit or the extrapolated pixels.
A standard regression model 
will fail in this situation, because
the training objective generally penalizes the distance between a \textit{single} prediction and the ground truth.
Faced with a problem like this, the best solution the regressor can produce is something which is in between the possibilities, since it minimizes the expected distance.
In the case of digits, this will most likely look like a meaningless blur that's an ``average image'' of all possible digits and all possible styles that could occur\footnote{
The denoising autoencoder~\cite{vincent2008extracting,bengio2013deep} can be seen as a slight generalization of the regression model, which might improve on its behavior.  
That is, we would say that the ``noise distribution'' simply deletes pixels, and so the denoising autoencoder must reconstruct the original image given the noisy version.  
Note, however, that this still doesn't solve the problem.  
The standard denoising autoencoder still requires that the conditional distribution of the original sample given the noisy sample follow a simple, parametric distribution.
This is not the case for complex data like image patches.
}.
What we need is an algorithm that takes in a string or an image, and produces a complex, multimodal distribution that we can sample from.
Enter the conditional variational autoencoder (CVAE)~\cite{sohn2015learning,walker2016uncertain}, which modifies the math in the previous section by simply conditioning the entire generative process on an input.
CVAEs allow us to tackle problems where the input-to-output mapping is one-to-many\footnote{Often called ``structured prediction'' in machine learning literature.}, without requiring us to explicitly specify the structure of the output distribution.

Given an input $X$ and an output $Y$, we want to create a model $P(Y|X)$ which maximizes the probability of the ground truth (I apologize for re-defining $X$ here.  However, standard machine learning notation maps $X$ to $Y$, so I will too).  
We define the model by introducing a latent variable $z\sim\mathcal{N}(0,I)$, such that:
\begin{equation}
 P(Y|X)=\mathcal{N}(f(z,X),\sigma^2*I)
\end{equation}
where $f$ is a deterministic function that we can learn from data.  We can rewrite Equations~\ref{eq:kl} through~\ref{eq:variational} conditioning on $X$ as follows:
\begin{equation}
    \mathcal{D}\left[Q(z|Y,X)\|P(z|Y,X)\right]=E_{z\sim Q(\cdot|Y,X)}\left[\log Q(z|Y,X) - \log P(z|Y,X) \right]
\end{equation}
\begin{equation}
\begin{array}{c}
    \mathcal{D}\left[Q(z|Y,X)\|P(z|Y,X)\right]=\hspace{20em}\\
    \hspace{1.5em}E_{z\sim Q(\cdot|Y,X)}\left[\log Q(z|Y,X) - \log P(Y|z,X) - \log P(z|X) \right] + \log P(Y|X)
\end{array}
\end{equation}
\begin{equation}
\begin{array}{c}
    \log P(Y|X) - \mathcal{D}\left[Q(z|Y,X)\|P(z|Y,X)\right]=\hspace{15em}\\
\hspace{8em}E_{z\sim Q(\cdot|Y,X)}\left[\log P(Y|z,X)  \right] - \mathcal{D}\left[Q(z|Y,X)\|P(z|X)\right]
\end{array}
\label{eq:cond_variational}
\end{equation}
Note that $P(z|X)$ is still $\mathcal{N}(0,I)$ because our model assumes $z$ is sampled independently of $X$ at test time.
The structure of this model is shown in Figure~\ref{fig:cvae}.

At test time, we can sample from the distribution $P(Y|X)$ by simply sampling $z\sim\mathcal{N}(0,I)$.  

\begin{figure*}
\begin{center}
\includegraphics[width=0.66\textwidth]{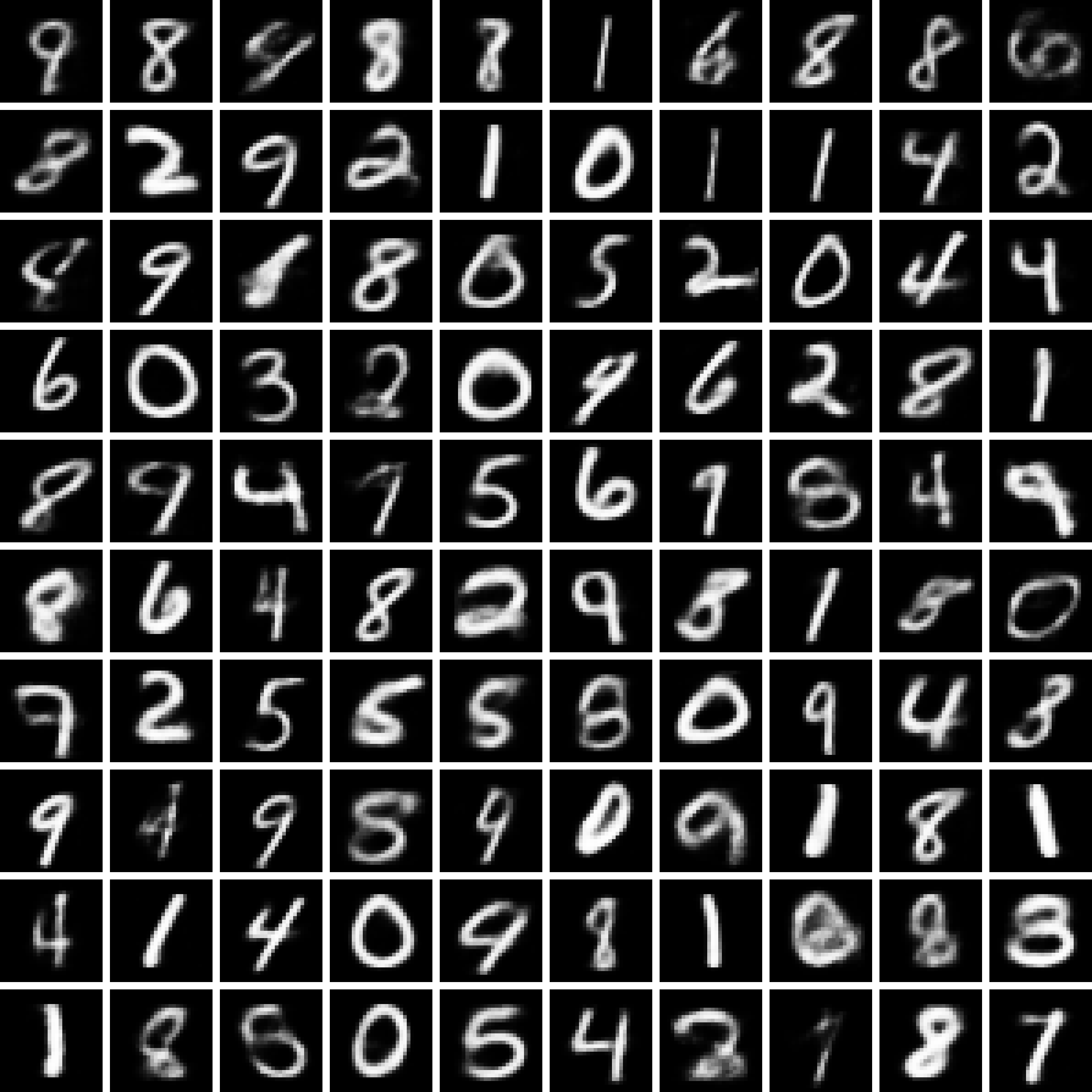} 
\end{center}
\caption{Samples from a VAE trained on MNIST.}
\label{fig:vae_out}
\end{figure*}

\section{Examples}
Implementations for these examples using Caffe~\cite{jia2014caffe} can be found online at:\\
\url{http://github.com/cdoersch/vae_tutorial}

\subsection{MNIST variational autoencoder}
\label{sec:mnist_vae}
To demonstrate the distribution learning capabilities of the VAE framework, let's train a variational autoencoder on MNIST.
To show that the framework isn't heavily dependent on initialization or network structure, we don't use existing published VAE network structures, but instead adapt the basic MNIST AutoEncoder example that's included with Caffe~\cite{jia2014caffe}. 
(However, we use ReLU non-linearities~\cite{krizhevsky2012imagenet} and ADAM~\cite{kingma2014adam}, since both are standard techniques to speed convergence.)  
Although MNIST is real-valued, it is constrained between 0 and 1, so we use the Sigmoid Cross Entropy loss for $P(X|z)$.
This has a probabilistic interpretation: imagine that we created a new datapoint $X^{\prime}$ by independently sampling each dimension as $X^{\prime}_i\sim \mbox{Bernoulli}(X_i)$.  
Cross entropy measures the \textit{expected} probability of $X^{\prime}$.  
Thus we're actually modeling $X^{\prime}$, the randomly binarized version of MNIST, but we're only giving $q$ a summary of this data $X$.  Admittedly this is not quite what the VAE framework prescribes but works well in practice, and is used in other VAE literature~\cite{Gregor15}.
Even though our model is considerably deeper than~\cite{Kingma14a} and \cite{Rezende14}, training the model was not difficult.
The training was run to completion exactly once (though the training was re-started the 5 times to find the learning rate which made the loss descend the fastest).
Digits generated from noise are shown in Figure~\ref{fig:vae_out}.
It's worth noting that these samples are difficult to evaluate since there's no simple way to measure how different these are from the training set~\cite{theis2015note}.
However, the failure cases are interesting: while most of the digits look quite realistic, a significant number are `in-between' different digits.
For example, the seventh digit from the top in the leftmost column is clearly in-between a 7 and a 9.
This happens because we are mapping a continuous distribution through a smooth function.

In practice, the model seems to be quite insensitive to the dimensionality of $z$, unless $z$ is excessively large or small. 
Too few $z$'s means the model can no longer capture all of the variation: less than 4 $z$ dimensions produced noticeably worse results.
Results with 1,000 $z$'s were good, but with 10,000 they were also degraded.  
In theory, if a model with $n$ $z$'s is good, then a model with $m>>n$ should not be worse, since the model can simply learn to ignore the extra dimensions.  
However, in practice, it seems stochastic gradient descent struggles to keep $\mathcal{D}[Q(z|X)||P(z)]$ low when $z$ is extremely large.  

\begin{figure*}
\begin{center}
\includegraphics[width=0.97\textwidth]{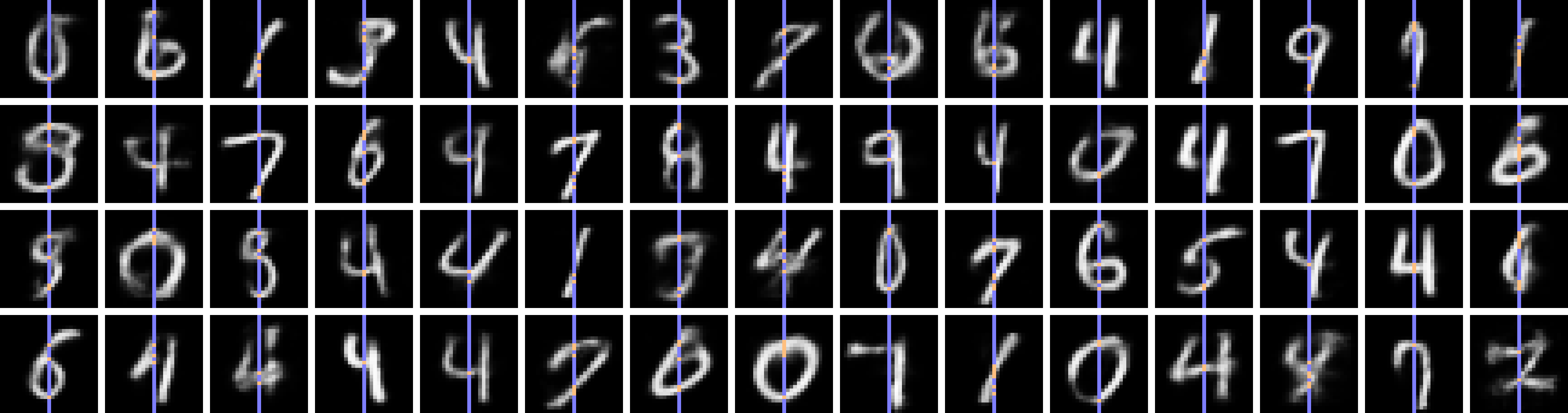} 
(a) CVAE
\includegraphics[width=0.97\textwidth]{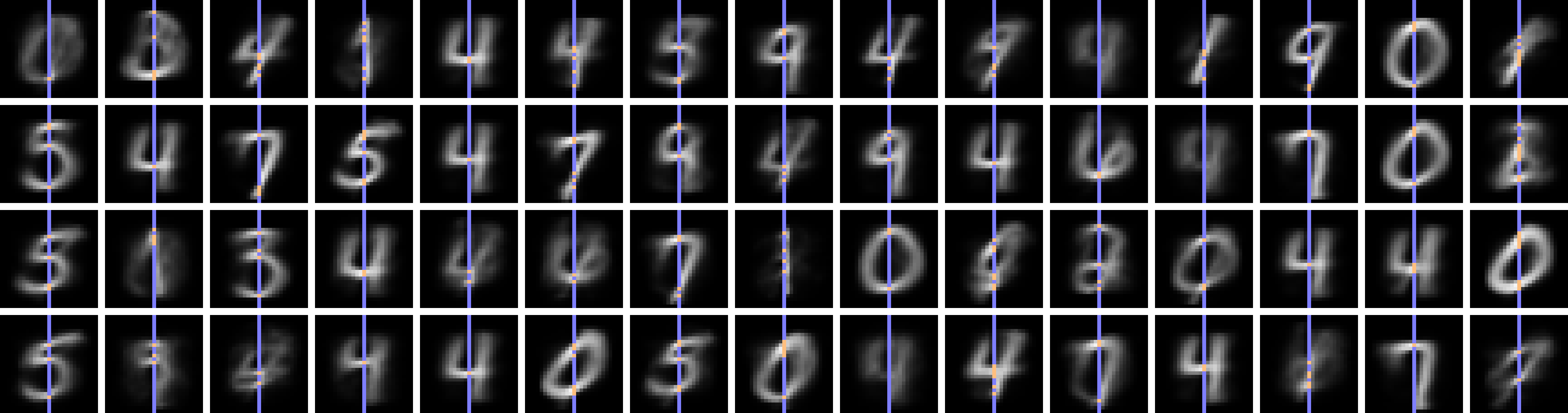} 
(b) Regressor
\includegraphics[width=0.97\textwidth]{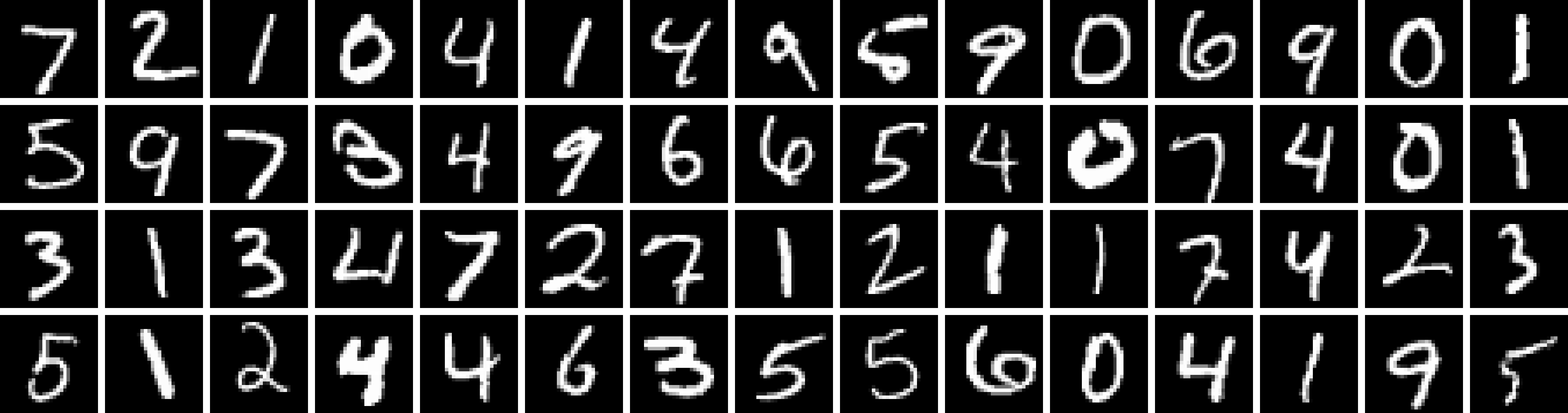} 
(c) Ground Truth
\end{center}
\caption{Samples from a CVAE trained on MNIST.  The input that the model is conditioning on is the central column, highlighted in blue and orange in the top two images.  The model must complete the digit given only these noisy binary values.  The three sets above are aligned spatially, so you can compare the generated images to the ground truth.}
\label{fig:cvae_out}
\end{figure*}

\section{MNIST conditional variational autoencoder}
I had originally intended to show a conditional variational autoencoder completing MNIST digits given only half of each digit. 
While a CVAE works quite well for this purpose, unfortunately a regressor actually works quite well also, producing relatively crisp samples.
The apparent reason is the size of MNIST.
A network with similar capacity to the one in section~\ref{sec:mnist_vae} can easily memorize the entire dataset, and so the regressor overfits badly.
Thus, at test time, it produces predictions that behave something like nearest-neighbor matching, which are actually quite sharp.
CVAE models are most likely to outperform simple regression when the output is ambiguous given a training example.
Therefore, let's make two modifications to the problem to make it more ambiguous, at the cost of making it somewhat more artificial.
First, the input is a single column of pixels taken from the middle of the digit.
In MNIST, each pixel has a value between 0 and 1, meaning that there is still enough information even in this single column of pixels for the network to identify a specific training example.
Therefore, the second modification is to replace each pixel in our column with a binary value (0 or 1), choosing 1 with probability equal to the pixel intensity.
These binary values were resampled each time a digit was passed to the network.
Figure~\ref{fig:cvae_out} shows the results.  
Note that the regressor model handles the ambiguity by blurring its output (although there are cases where the regressor is suspiciously confident when making wrong guesses, suggesting overfitting is still an issue). 
The blur in the regressor's output minimizes the distance to the set of many digits which might have produced the input. 
The CVAE, on the other hand, generally picks a specific digit to output and does so without blur, resulting in more believable images.

\textbf{Acknowledgements:} 
Thanks to everyone in the UCB CS294 Visual Object And Activity Recognition group and CMU Misc-Read group, and to many others who encouraged me to convert the presentation I gave there into a tutorial.
I would especially like to thank Philipp Kr{\"a}henb{\"u}hl, Jacob Walker, and Deepak Pathak for helping me formulate and refine my description of the method, and Kenny Marino for help editing.
I also thank Abhinav Gupta and Alexei Efros for helpful discussions and support, and Google for their fellowship supporting my research.

\appendix
\section{Proof in 1D that VAEs have zero approximation error given arbitrarily powerful learners.}
\label{appendix:convergence}
Let $P_{gt}(X)$ be a 1D distribution that we are trying to approximate using a VAE.
We assume that $P_{gt}(X)>0$ everywhere, that it is infinitely differentiable, and all the derivatives are bounded.
Recall that a variational autoencoder optimizes
\begin{equation}
 \log P_\sigma(X)-\mathcal{D}[Q_\sigma(z|X)\|P_\sigma(z|X)]
\end{equation}
where $P_\sigma(X|z)=\mathcal{N}(X|f(z),\sigma^2)$ for $z\sim \mathcal{N}(0,1)$, $P_\sigma(X)=\int_z P_\sigma(X|z) P(z) dz$, and $Q_\sigma(z|X)=\mathcal{N}(z|\mu_\sigma(X),\Sigma_\sigma(X))$.
We make the dependence on $\sigma$ explicit here since we will send it to 0 to prove convergence.
The theoretical best possible solution is where $P_\sigma=P_{gt}$ and $\mathcal{D}[Q_\sigma(z|X)\|P_\sigma(z|X)]=0$.
By ``arbitrarily powerful'' learners, we mean that if there exist $f$, $\mu$ and $\Sigma$ which achieve this best possible solution, then the learning algorithm will find them.
Hence, we must merely show that such an $f$, $\mu_\sigma$, and $\Sigma_\sigma$ exist.
First off, $P_{gt}$ can actually be described arbitrarily well as $P_{gt}(X)=\int_z\mathcal{N}(X|f(z),\sigma^2) P(z) dz$ as $\sigma$ approaches 0.  
To show this, let $F$ be the cumulative distribution function (CDF) of $P_{gt}$, and let $G$ be the CDF of $\mathcal{N}(0,1)$, which are both guaranteed to exist.  
Then $G(z)$ is distributed $Unif(0,1)$ (the uniform distribution), and therefore $f(z)=F^{-1}(G(z))$ is distributed $P_{gt}(X)$.  
This means that as $\sigma\rightarrow 0$, the distribution $P(X)$ converges to $P_{gt}$.

From here, we must simply show that $\mathcal{D}[Q_\sigma(z|X)\|P_\sigma(z|X)]\rightarrow 0$ as $\sigma \rightarrow 0$.
Let $g(X)=G^{-1}(F(X))$, i.e., the inverse of $f$, and let $Q_\sigma(z|X)=\mathcal{N}(z|g(X),(g^{\prime}(X)*\sigma)^2)$.
Note that $\mathcal{D}[Q_\sigma(z|X)||P_\sigma(z|X)]$ is invariant to affine transformations of the sample space.
Hence, let $Q^0(z^0|X)=\mathcal{N}(z^0|g(X),g^{\prime}(X)^2)$ and $P^0_\sigma(z^0|X)=P_\sigma(z=g(X)+(z^0-g(X))*\sigma|X)*\sigma$.
When I write $P(z=...)$, I am using the PDF of $z$ as a function, and evaluating it at some point.
Then:
\begin{equation}
 \mathcal{D}[Q_\sigma(z|X)\|P_\sigma(z|X)]=\mathcal{D}[Q^0(z^0|X)\|P^0_\sigma(z^0|X)]
\end{equation}
where $Q^0(z^0|X)$ does not depend on $\sigma$, and its standard deviation is greater than $0$.
Therefore, it is sufficient to show that $P^0_\sigma(z^0|X)\rightarrow Q^0(z^0|X)$ for all $z$.
Let $r=g(X)+(z^0-g(X))*\sigma$.
Then:
\begin{equation}
\begin{array}{rl}
  P^{0}_\sigma(z^0|X) & =P_\sigma(z=r|X=X)*\sigma \vspace{1em}\\
  &=\dfrac{P_\sigma(X=X|z=r)*P(z=r)*\sigma}{P_\sigma(X=X)}.
\end{array}
\end{equation}
Here, $P_\sigma(X) \rightarrow P_{gt}(X)$ as $\sigma \rightarrow 0$, which is a constant, and $r\rightarrow g(X)$ as $\sigma \rightarrow 0$, so $P(r)$ also tends to a constant.
Together with $\sigma$, they ensure that the whole distribution normalizes.
We will wrap them both in the constant $C$.
\begin{equation}
  =C*\mathcal{N}(X|f(g(X)+(z^0-g(X))*\sigma),\sigma^{2}).
\end{equation}
Next, we do a Taylor expansion of $f$ around $g(X)$:
\begin{equation}
  =C*\mathcal{N}\left(X\middle| X+f^{\prime}(g(X))*(z^0-g(X))*\sigma+\sum_{n=2}^{\infty}\frac{f^{(n)}(g(X))((z^0-g(X))*\sigma)^n}{n!},\sigma^{2}\right).
\end{equation}
Note that $\mathcal{N}(X|\mu,\sigma)=(\sqrt{2\pi}\sigma)^{-1}\exp\left(\frac{-(x-\mu)^2}{2\sigma^2}\right)$.  We rewrite the above using this formula, rearrange terms, and re-write the result as a Gaussian to obtain:
\begin{equation}
  =\frac{C*f^{\prime}(g(X))}{\sigma}*\mathcal{N}\left(z^0\middle|g(X)-\sum_{n=2}^{\infty}\frac{f^{(n)}(g(X))((z^0-g(X))*\sigma)^n}{n!*f^{\prime}(g(X))*\sigma},\frac{1}{f^{\prime}(g(X))^2}\right).
  \label{eq:taylor}
\end{equation}
Note $1/f^{\prime}(g(X))=g^\prime(X)$, since $f=g^{-1}$.  
Furthermore, since $f^{(n)}$ is bounded for all $n$, all terms in the sum tend to 0 as $\sigma\rightarrow 0$.  
$C$ must make the distribution normalize, so we have that the above expression:
\begin{equation}
  \rightarrow\mathcal{N}\left(z^0|g(X),g^\prime(X)^2\right)=Q^{0}(z^0|X) \qed
\end{equation}
Looking at Equation~\ref{eq:taylor}, the bulk of the approximation error in this setup comes from the curvature of $g$, which is mostly determined by the curvature of the c.d.f. of the ground truth distribution.



\bibliographystyle{unsrt}

\bibliography{vae_tutorial}


\end{document}